%% file: latex/main.tex
\documentclass[11pt]{article}

\input{latex/commands}

\title{\tool: AST-Guided Translation of Natural Language into First-Order Logic with Large Language Models}

\author{
  \textbf{Rizky Ramadhana Putra}\textsuperscript{1}, 
  \textbf{Raihan Sultan Pasha Basuki}\textsuperscript{2}, 
  \textbf{Yutong Cheng}\textsuperscript{1}, 
  \textbf{Peng Gao}\textsuperscript{1} \\
  \\
  \textsuperscript{1}Virginia Tech, Blacksburg, VA, USA \\
  \textsuperscript{2}Universitas Ary Ginanjar, Jakarta, Indonesia \\
  \texttt{\{rizky, yutongcheng, penggao\}@vt.edu} \\
  \texttt{raihansultan.pashabasuki@students.uag.ac.id}
}

\begin{document}
\maketitle

\input{latex/sections/abstract}

% ==============================================================

\section{Introduction} \label{sec:intro}

\input{latex/sections/introduction}

% ==============================================================

\section{Related Work} \label{sec:related_works}

\input{latex/sections/related_works}

% ==============================================================

% \section{Dataset Generation} \label{sec:dataset}

% \input{latex/sections/dataset}

% ==============================================================
\section{Preliminary} \label{sec:preliminary}

\input{latex/sections/preliminary}

\section{\tool Design} \label{sec:design}

\input{latex/sections/design}

% ==============================================================

\section{Evaluation} \label{sec:eval}

\input{latex/sections/evaluation}

% ==============================================================

\section{Conclusion} \label{sec:conclusion}

\input{latex/sections/conclusion}

% ==============================================================

\section*{Ethical Considerations} \label{sec:ethics}

\input{latex/sections/ethics.tex}

% ==============================================================

% \newpage

\section*{Limitations} \label{sec:limitations}

\input{latex/sections/limitations.tex}

% ==============================================================

\section*{Acknowledgments} \label{acknowledgement}

\input{latex/sections/acknowledgement}

% ==============================================================
\bibliography{custom}
\clearpage
\appendix

\section{System Prompts} \label{sec:system_prompts}

\input{latex/sections/appendix}

\end{document}

%% file: latex/commands.tex
\usepackage[final]{acl}

\usepackage{times}
\usepackage[dvipsnames]{xcolor}
\usepackage{xcolor}
\usepackage{latexsym}

\usepackage[T1]{fontenc}
\usepackage[utf8]{inputenc}

\usepackage{microtype}

\usepackage{inconsolata}

\usepackage{graphicx}

\usepackage{tcolorbox}
\tcbuselibrary{listings,skins,breakable}
\usepackage{caption}
\usepackage{tcolorbox}

\usepackage{enumitem}

\usepackage{xspace}
\usepackage{amsmath}

\usepackage[capitalize,nameinlink]{cleveref}
\usepackage{subcaption}

\def\circled#1{%
  \hspace{0.2em}
  #1%
  \pdfliteral{
    q .5 w
    10 0 0 10 -2.5 3.5 cm .05 w .5 0 m
    .5 .276 .276 .5 0 .5 c -.276 .5 -.5 .276 -.5 0 c
    -.5 -.276 -.276 -.5 0 -.5 c .276 -.5 .5 -.276 .5 0 c h
    S Q
  }%
  \hspace{0.2em}
}

\usepackage{soul}

\newcommand{\hlc}[2][yellow]{{%
    \colorlet{foo}{#1}%
    \sethlcolor{foo}\hl{#2}}%
}

\usepackage{mdframed}
\usepackage{tikz}
\usepackage{xcolor}
\usepackage{algorithm}
\usepackage[spaceRequire=false]{algpseudocodex}
\usepackage{makecell}
\usepackage{multirow}

\newcommand{\dashedbox}[2][black]{%
  \tikz[baseline] \node[
    draw=#1,
    dash pattern=on 2pt off 2pt,
    line cap=round,
    line width=1pt,
    inner xsep=2pt,
    inner ysep=0.4pt,
    outer sep=0pt,   % <-- prevents extra spacing
    anchor=base
  ]{#2};\xspace
}

\crefname{figure}{Fig.}{Figs.}
\crefname{listing}{Listing}{Listings}
\crefname{section}{Section}{Sections}
\crefname{table}{Table}{Tables}
\crefname{algorithm}{Algorithm}{Algorithms}
\crefname{line}{Line}{Lines}
\crefname{equation}{Eq.}{Eqs.}
\crefname{rq}{RQ}{RQs}

\renewcommand{\paragraph}[1]{\smallskip\noindent\textbf{#1.}}

\newif\ifshowcomment
\showcommenttrue
\showcommentfalse

\ifshowcomment
    \newcommand{\cicip}[1] {{\footnotesize\color{Aquamarine}[CiCi: #1]\newline}}
    \newcommand{\cici}[1] {{\footnotesize\color{Aquamarine}[CiCi: #1]}}
    \newcommand{\pgao}[1] {{\footnotesize\color{red}[Peng: #1]}}
    \newcommand{\dawn}[1] {{\footnotesize\color{blue}[Dawn: #1]}}
    
    \newcommand{\rizky}[1] {{\footnotesize\color{purple}[Rizky: #1]}}
    \newcommand{\raihan}[1] {{\footnotesize\color{ForestGreen}[Raihan: #1]}}

\else
    \newcommand{\cici}[1]{}
    \newcommand{\cicip}[1]{}
    \newcommand{\pgao}[1]{}
    \newcommand{\rizky}[1]{}
    \newcommand{\raihan}[1]{}
    \newcommand{\dawn}[1]{}

\fi

\newcommand{\tool}{\textsc{NL2Logic}\xspace}

\newcommand{\parser}{\emph{semantic parser}\xspace}
\newcommand{\generator}{\emph{AST-guided generator}\xspace}

%% file: latex/sections/abstract.tex
\begin{abstract}
Automated reasoning is critical in domains such as law and governance, where verifying claims against facts in documents requires both accuracy and interpretability.
Recent work has adopted a structured reasoning paradigm that parses first-order logic (FOL) rules from natural language and delegates inference to automated solvers.
With the rise of large language models (LLMs), methods such as GCD and CODE4LOGIC leverage their reasoning and code generation capabilities to enhance logic parsing.
However, these approaches suffer from (1) fragile syntax control, due to weak enforcement of global grammar consistency, and (2) low semantic faithfulness, as they lack fine-grained clause-level semantic understanding.
To address these challenges, we propose \emph{\tool}, a FOL translation framework that uses an AST as an intermediate layer, combining a recursive LLM-based semantic parser with an AST-guided generator that deterministically produces solver-ready code.
On the FOLIO, LogicNLI, and ProofWriter benchmarks, \tool attains 99\% syntactic accuracy and improves semantic correctness by 30\% over state-of-the-art baselines.
Moreover, integrating \tool into Logic-LM yields near-perfect executability and improves downstream reasoning accuracy by ~31\% over Logic-LM’s original few-shot unconstrained FOL translation module.

\end{abstract}

%% file: latex/sections/introduction.tex
Natural language documents, especially in domains such as law, policy, and governance, encode complex logical relations that must be interpreted precisely for downstream reasoning and compliance checking. However, natural language is inherently ambiguous and unstructured, making it difficult for machines, and even humans, to ensure logical consistency, detect contradictions, or verify claims across documents. This gap has motivated research on \emph{automated reasoning over natural language}, where models assess whether a claim is entailed by supporting evidence. Early approaches rely on neural entailment frameworks \cite{evans2018can, bowman-etal-2015-recursive, rocktaschel2016reasoning} that employ neural networks to classify entailment and contradiction. However, these models remain opaque black boxes, lacking interpretability and formal verifiability. To improve transparency and explainability, recent research introduces a structured reasoning paradigm that first \emph{derives explicit logical representations from text} and then verifies claims through automated reasoning engines. This transition improves inference transparency and auditability by explicitly representing intermediate proof steps, rather than only final results.

In recent years, Large Language Models (LLMs) have shown notable progress in logical reasoning, particularly when guided by prompting strategies such as few-shot examples and Chain-of-Thought (CoT) prompting. Building on this progress, recent research has developed \emph{LLM-powered approaches} for translating natural language (NL) into first-order logic (FOL), leveraging the models’ strong natural language understanding and code generation capabilities to enable automated reasoning with explicit semantics such as FOL.

To improve NL-to-FOL translation performance, several approaches have developed specialized pipelines that enhance LLMs with additional structural guidance and symbolic control.
A representative work, Grammar-Constrained Decoding (GCD)~\cite{gcd}, enforces token-level syntactic correctness by encoding the target formal language as a context-free grammar
and constraining the LLM’s decoding process to adhere to this grammar.
CODE4LOGIC~\cite{code4logic} adopts a complementary approach.
It leverages the code understanding capability of code generation models.
Rather than generating FOL formulas directly, it uses few-shot prompting to produce Python code that encodes the grammar tree, which, when executed, generates the corresponding FOL rule.

Despite their contributions, these methods face two key limitations:

\begin{itemize}[leftmargin=*]
    \item \textbf{Fragile syntax control.} 
    GCD restricts generation at the token level through context-free grammars. While this enforces local syntactic correctness, grammatically valid outputs do not necessarily constitute valid logical statements, such as those requiring variable or signature consistency.
    CODE4LOGIC uses in-context learning (ICL) to generate a grammar tree and guides subsequent code generation with this structure. 
    However, despite leveraging ICL, the free-form generation process remains prone to hallucination, and any errors in the tree can propagate into the following code generation stage, leading to invalid or inconsistent outputs.

    \item \textbf{Limited semantic faithfulness.} 
    Existing approaches formulate NL–to–FOL translation as a single-step text-to-text task, mapping entire paragraphs to complete logical programs and forcing models to perform comprehension and translation simultaneously.
    Handling each clause iteratively is essential, as clauses often encode complex logical relations that can overwhelm the LLM if processed and translated all at once.
    Without this iterative decomposition, the LLM tends to rely on shallow token prediction rather than genuine logical understanding, making hallucinations more likely when sentence structures become complex.
\end{itemize}

To address these limitations, we propose \tool, a framework that translates natural language sentences into \emph{semantically faithful} and \emph{syntactically correct} logical rules.
Rather than adopting an imprecise, unconstrained, and single-step approach, the \parser decomposes each sentence into clauses and iteratively extracting first-order logic components (e.g., predicates, quantifiers, and logical connectives) to construct a First-Order Logic Abstract Syntax Tree (FOLAST).
Parsing each clause iteratively enables the model to make controlled, grammar-guided decisions, ensuring clause-level accuracy that composes into a globally coherent logical representation.

Then, the \generator ensures \textbf{syntactic correctness} by deterministically compiling the FOLAST through a two-pass algorithm: the first pass registers all constants, variables, and relation signatures, while the second pass assembles scoped expressions following solver-specific grammar.
This design enforces strict syntactic validity while preserving semantic alignment, producing executable logical rules compatible with solvers such as Z3 \cite{z3} and SMT-LIB \cite{smt}.

We evaluate \tool on three widely used natural language inference (NLI) datasets, FOLIO~\cite{folio}, ProofWriter~\cite{proofwriter}, and LogicNLI~\cite{logicnli}, using the Z3 reasoning engine for consistent formal verification.
We also integrate \tool into Logic-LM~\cite{logiclm}, a representative neuro-symbolic framework, demonstrating its practical value as a modular component.

The evaluation addresses three research questions: (RQ1) whether the generated formulas are syntactically valid; (RQ2) whether they preserve the intended semantics for entailment prediction;
(RQ3) whether the integration of \tool improves existing neuro-symbolic systems.
Across twelve models ranging from 0.5B to 27B parameters, \tool achieves near-perfect syntax correctness, improves semantic accuracy by an average of 30\% over the strongest baseline (GCD), and improves the existing neuro-symbolic framework by 31\% on downstream reasoning task over Logic-LM’s original unconstrained FOL translation module.
These results demonstrate that our decoupled, AST-based design provides stronger syntactic control and more faithful semantic alignment, advancing automated reasoning over natural language through formal, interpretable symbolic representations.

To facilitate future research, we release our implementation at: \url{https://github.com/peng-gao-lab/nl2logic}.

%% file: latex/sections/related_works.tex
\paragraph{Natural Language to Formal Logic Translation}
Research on formal logic translation has evolved across several paradigms. Early rule-based methods~\cite{bos2005recognising, zettlemoyer2012learning, barker2009dimensions, abzianidze2017langpro} offer precise control but lack robustness to linguistic variation. Neural models later generated logical forms directly from text~\cite{lu2022parsing, cao2019semantic}, improving generalization but struggling with rare operators and complex syntax. More recently, neuro-symbolic systems commonly adopt unconstrained few-shot LLM translation integrated with logic solvers~\cite{logiclm,linc,symbcot,fairr,explanation_refiner,faithful_refiner}.
Structured pipelines that add syntactic constraints~\cite{gcd, code4logic} further improve performance, yet the resulting formulas often remain semantically unfaithful to the source text.

\begin{figure*}[]
    \centering
    \includegraphics[]{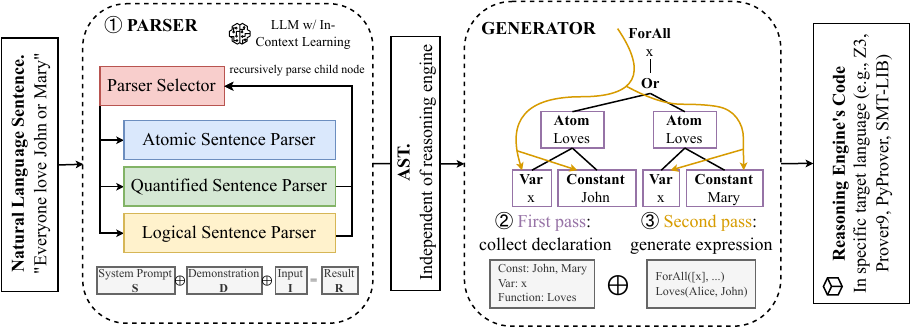}
    \caption{Overview of \tool. The semantic parser (\circled{1}) takes a natural language sentence and outputs a first-order logic abstract syntax tree (FOLAST) through a recursive, top-down approach. The AST is then compiled into the reasoning engine's target language through a two-pass algorithm. The first pass (\circled{2}) collects constant, variable, and predicate signature declarations. The second pass (\circled{3}) generates the concrete logical expression.}
    \label{fig:overview}
\end{figure*}

\paragraph{LLMs for Logical Reasoning}
The development of logical reasoning capabilities in LLMs has seen significant progress through a range of approaches. One line of work breaks down complex reasoning into intermediate steps, often referred to as chain-of-thought prompting~\cite{wei2022chain}, while others show that simple step-by-step prompting can yield similar benefits without explicit examples ~\cite{kojima2022large}.
However, these approaches lack the formal proof.
To address this, some frameworks combine LLMs with automated reasoning tools to improve faithfulness~\cite{creswell2205selection}.
Recent research has further explored integrating LLMs with symbolic solvers~\cite{wang2023grammar}, treating LLMs as logical parsers rather than independent reasoners.
In summary, prior work either lacks formal proof for logical reasoning or produces unfaithful logical forms due to overly simple, unconstrained translation methods.

%% file: latex/sections/preliminary.tex
\paragraph{First-Order Logic}
First-Order Logic (FOL) is a formal system for expressing statements about entities, their properties, and relations \cite{fol}. Its syntax comprises \emph{constants}, \emph{variables}, \emph{functions}, \emph{relations}, \emph{quantifiers} (e.g., $\forall$, $\exists$), and \emph{logical connectives} (e.g., $\land$, $\lor$, $\rightarrow$, $\lnot$). A well-formed FOL formula such as $\forall x.\, \textit{Human}(x) \rightarrow \textit{Mortal}(x)$ captures meaning in a precise, machine-verifiable form. Unlike natural language, FOL enforces strict grammar, enabling unambiguous interpretation and compatibility with automated reasoning systems.

\begin{figure}[t]
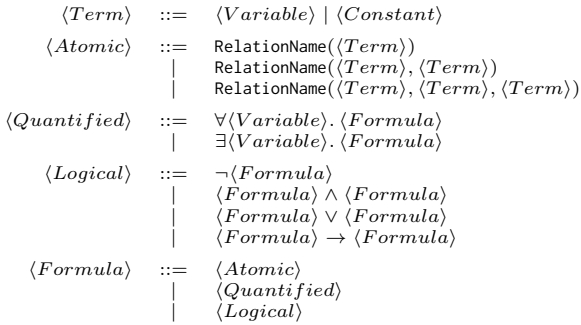

\centering
\scriptsize
\[
\begin{array}{rcl}
\langle Term \rangle      &::=& \langle Variable \rangle \mid \langle Constant \rangle \\[4pt]

\langle Atomic \rangle &::=& \texttt{RelationName}(\langle Term \rangle) \\ 
                          &\mid&\texttt{RelationName}(\langle Term \rangle,\langle Term \rangle) \\
                          &\mid& \texttt{RelationName}(\langle Term \rangle,\langle Term \rangle,\langle Term \rangle) \\[4pt]

\langle Quantified \rangle &::=& \forall \langle Variable \rangle.\,\langle Formula \rangle \\
                          &\mid& \exists \langle Variable \rangle.\,\langle Formula \rangle \\[4pt]

\langle Logical \rangle &::=& \neg \langle Formula \rangle \\
                          &\mid& \langle Formula \rangle \land \langle Formula \rangle \\
                          &\mid& \langle Formula \rangle \lor \langle Formula \rangle \\
                          &\mid& \langle Formula \rangle \rightarrow \langle Formula \rangle \\[4pt]

\langle Formula \rangle   &::=& \langle Atomic \rangle \\ 
                          &\mid& \langle Quantified \rangle \\
                          &\mid& \langle Logical \rangle \\
\end{array}
\]
\caption{Formal notation of abstract syntax tree (AST)}
\label{fig:grammar}
\vspace{-1ex}
\end{figure}

\paragraph{Abstract Syntax Tree}
An Abstract Syntax Tree (AST) is a tree-based data structure, commonly used in compilers and interpreters to represent the syntactic structure of a program \cite{ast}.
Each node in an AST corresponds to a syntactic construct (e.g., operator, expression, declaration), enabling structural analysis and systematic code generation.
We extend this concept to a first-order logic AST (\emph{FOLAST}) defined by the standard FOL grammar in \cref{fig:grammar}, serving as a backend-independent intermediate representation linking natural language to the generated code that will be executed by automated reasoning engines.

\paragraph{Reasoning Engines and Target Languages}
Automated reasoning engines such as Z3~\cite{z3}, Prover9~\cite{prover9}, and SMT-LIB~\cite{smt} solvers evaluate FOL statements for satisfiability, entailment, and consistency. Each engine requires strict, engine-specific syntax, e.g., Z3’s Python API or SMT-LIB’s symbolic format. However, these syntactic forms are rarely observed during LLM pretraining.
Hence, directly generating solver code in an end-to-end manner leads to high syntax errors and inconsistent logical structures.
Our \emph{key insight} is to decouple logical parsing from code generation by introducing an intermediate AST representation, which captures the logical structure in an engine-agnostic form and can then be deterministically compiled into the target reasoning language.
This separation enforces syntactic correctness, facilitates multi-engine generalization, and avoids LLM hallucinations tied to solver-specific syntax.

%% file: latex/sections/design.tex
\tool adopts a parser-generator architecture designed to ensure syntactic correctness and semantic faithfulness in translating natural language into formal logic. 
As illustrated in \cref{fig:overview}, the pipeline consists of two stages: a \parser, which converts natural language text into a structured first-order logic abstract syntax tree (FOLAST), defined by the standard grammar in \cref{fig:grammar}, and an \generator, which compiles the FOLAST into solver-executable code (e.g., Z3, SMT-LIB). We next describe each component.

\subsection{Preprocessing}

Documents are divided into well-defined sentences leveraging prior work on sentence boundary detection, as directly feeding long paragraphs into the parser risks hallucination and error propagation. 
Instead of rule-based splitting that relies only on punctuation and heuristics, we adopt SaT~\cite{sat,wtpsplit}, a learning-based model that predicts sentence boundaries using contextual and lexical cues.
This ML-based approach distinguishes true sentence endings from punctuation used in abbreviations (e.g., \textit{U.S.}, \textit{Prof.}) or numeric expressions, thereby avoiding fragmentation errors. 
As a result, each logical sentence is reliably isolated, providing clean and accurate input to the semantic parser.

\subsection{Semantic Parser}

The semantic parser is the first and most critical stage of \tool. It maps natural language sentences into a first-order logic abstract syntax tree (FOLAST; described in \cref{sec:preliminary}) that strictly enforces logical grammar. Without this stage, directly prompting an LLM to generate FOL symbols or solver-specific code (e.g., Z3, SMT-LIB) often yields syntax errors, hallucinations, or undeclared variables.
This is demonstrated in our evaluations (\cref{sec:eval}) and caused by such formats are rarely observed during pre-training. 
By isolating parsing as a dedicated component, we guarantee that natural language is first converted into a syntactically valid and semantically transparent representation. 

\begin{figure}[!t]
\centering

    \begin{subfigure}{\linewidth}
        \begin{mdframed}
            \small
            Rina is either a \hlc[cyan!30]{student who is unaware that caffeine is a drug}, \hlc[pink]{or} \hlc[cyan!30]{she is not a student and is aware that caffeine is a drug}
        \end{mdframed}
        \vspace{-1ex}
        \caption{The parser decomposes a sentence containing a logical operator into its \hlc[pink]{operator} and \hlc[cyan!30]{operand(s)}.}
    \end{subfigure}
    
    \vspace{1ex}
    
    \begin{subfigure}{\linewidth}
        \begin{mdframed}
                \small
                - Rina is a student who is unaware that caffeine is a drug
                
                - Rina is not a student and is aware that caffeine is a drug
        \end{mdframed}
        \vspace{-1ex}
        \caption{The extracted operands are rewritten as standalone sentences. Since they are not necessarily atomic, each operand is recursively fed back to the parser.}
    \end{subfigure}

    \vspace{0ex}

    \begin{subfigure}{\linewidth}
        \centering
        \includegraphics[]{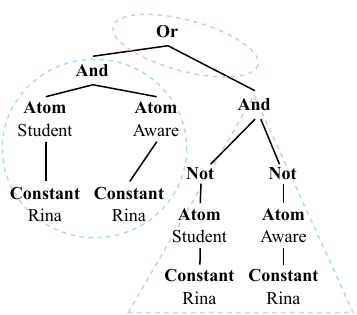}
         \caption{The complete AST representation.
        The parser identifies only the \dashedbox[pink]{outermost structure}, while the \dashedbox[cyan!30]{operands} are recursively fed back to the parser.}
    \end{subfigure}

    \vspace{1ex}

    \begin{subfigure}{\linewidth}
        \begin{mdframed}
                \small
                \centering
                $(\text{Student}(\text{Rina}) \wedge \text{Aware}(\text{Rina})) \vee (\neg\text{Student}(\text{Rina}) \wedge \neg\text{Aware}(\text{Rina}))$
        \end{mdframed}
        \vspace{-1ex}
        \caption{The first-order logic rule representation}
    \end{subfigure}

\caption{\texttt{LogicalSentenceParser} example.}
\label{fig:operator-example}
\vspace{-1ex}
\end{figure}

Conventional NLP parsers (e.g., dependency or constituency parsers) are inadequate for translating NL into formal logic.
Rule-based parsers \cite{rbp1,rbp2} fail to capture the diversity and ambiguity of natural sentences, while conventional ML-based parsers \cite{mlp1,mlp2} require large annotated corpora of NL and FOL pairs.
In contrast, large language models (LLMs) can perform in-context learning \cite{icl},
allowing them to follow explicit grammar constraints and generate structured FOLAST outputs.
Therefore, an \emph{LLM-based parser} is both necessary and practical: it combines pretrained linguistic knowledge with these formal grammar constraints to produce accurate and generalizable logical representations.

Our parser operates recursively through specialized sub-modules, as shown in \cref{fig:overview}.
The \emph{Parser Selector} first classifies each sentence as \texttt{atomic}, \texttt{quantified}, or \texttt{logical}.
An \texttt{atomic} sentence expresses a single relation, a \texttt{quantified} sentence introduces a quantifier (e.g., $\forall$, $\exists$), and a \texttt{logical} sentence contains logical connectives (e.g., $\land$, $\lor$, $\rightarrow$, $\neg$).
Then, the corresponding sub-parser is invoked: the \emph{Atomic Sentence Parser} extracts predicates and arguments; the \emph{Quantified Sentence Parser} identifies quantifiers and variables, then recursively parses the quantified scope; and the \emph{Logical Sentence Parser} detects operators, splits the input into operands, and recursively processes each operand.
At each step, the parser focuses only on the outermost construct while delegating the remaining sub-sentences to recursive parsing calls.
This \emph{top-down recursion} incrementally builds the AST, mirroring the hierarchical composition of FOL expressions.
This design limits error propagation, reduces the LLM’s cognitive load, and maintains semantic faithfulness across sentences of varying complexity.
The complete system prompts for each parser are provided in \cref{sec:system_prompts}.

\begin{figure}[!t]
\centering

    \begin{subfigure}{\linewidth}
        \begin{mdframed}
            \small
            \hlc[pink]{All people} \hlc[cyan!30]{who regularly drink coffee are dependent on caffeine}.
        \end{mdframed}
        \vspace{-1ex}
        \caption{The parser decomposes the sentence into its \hlc[pink]{quantifier} and \hlc[cyan!30]{scope}.}
    \end{subfigure}
    
    \vspace{1ex}
    \begin{subfigure}{\linewidth}
        \begin{mdframed}
            \small
            x that regularly drink coffee are dependent on caffeine
        \end{mdframed}
        \vspace{-1ex}
        
        \caption{The scope is rewritten such that the quantified subject is replaced by a variable. Since the scope is not necessarily atomic, it is recursively fed back into the parser.}
    \end{subfigure}
    
    \vspace{1ex}
    \begin{subfigure}{\linewidth}
        \centering
        \includegraphics[]{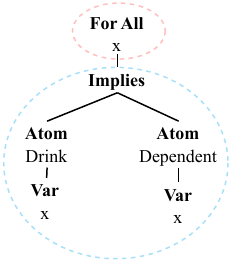}
        \caption{The complete AST representation.
        The parser identifies only the \dashedbox[pink]{outermost structure}, while the \dashedbox[cyan!30]{scope} is recursively fed back to the parser.}
    \end{subfigure}

    \vspace{1ex}
    \begin{subfigure}{\linewidth}
        \begin{mdframed}
                \small
                \centering
                $\forall x \, (\text{Drink}(x) \rightarrow \text{Dependent}(x))$
        \end{mdframed}
        \vspace{-1ex}
        \caption{The first-order logic rule representation.}
    \end{subfigure}

\caption{\texttt{QuantifiedSentenceParser} example.}
\label{fig:quantifier-example}
\vspace{-1ex}

\end{figure}

To illustrate the recursive mechanism, \cref{fig:operator-example} and \cref{fig:quantifier-example} present two representative cases, compound and quantified sentences.
\cref{fig:operator-example} shows how a disjunctive sentence is parsed by identifying the outermost operator (\texttt{Or}) and recursively decomposing its operands, each containing internal conjunctions and negations.
At each step, the parser enforces local correctness by validating the syntactic and semantic consistency of each node before integrating it into the higher-level tree.
For example, it ensures that every predicate has a valid argument (e.g., \texttt{Student(Rina)}) and that logical connectives such as \texttt{And} or \texttt{Not} combine clauses rather than partial phrases.
\cref{fig:quantifier-example} then illustrates a quantified structure, where the parser isolates the quantifier (\texttt{ForAll}), abstracts the variable (\texttt{x}), and recursively parses its scoped clause into atomic predicates such as \texttt{Drink(x)} and \texttt{Dependent(x)}.
These two examples show how our parser consistently handles different logical forms by decomposing them into their outermost constructs and deferring sub-sentences to recursive parsing.

\subsection{AST-Guided Generator}

The AST-guided generator converts the FOLAST produced by the parser into solver-ready code (e.g., Z3 or SMT-LIB). 
The generator deterministically maps each AST node to target-language expressions, mimicking how source-to-source compilers work (e.g., Cython \cite{cython}).
Its workflow follows a \emph{two-pass algorithm}: the first pass traverses the AST to collect all variable, constant, and relation signature declarations, establishing a consistent global context; the second pass revisits each node to emit logical expressions that respect operator precedence, quantifier scope, and relation arity.
The generator produces executable code in the reasoning engine’s target language, preserving syntactic consistency and faithfully reflecting the logical structure defined by the FOLAST.
We describe each pass below.

\begin{algorithm}[!t]
    \caption{First pass to collect declaration in an AST}
    \label{alg:first_pass}
    \begin{algorithmic}
        \Require $N$ \Comment{set of nodes in an AST}
        \Ensure $D$ \Comment{set of declaration}
        \State $D \gets \emptyset$
        \For{$n \in N$} 
            \If {$n$ is a variable}
                \State $D \gets D \cup \{\Call{DeclareVar}{n}\}$
            \ElsIf {$n$ is a constant}
                \State $D \gets D \cup \{\Call{DeclareConst}{n}\}$
            \ElsIf {$n$ is a relation}
                \State $D \gets D \cup \{\Call{DeclareRelation}{n}\}$
            \Else\; continue
            \EndIf
        \EndFor
    \end{algorithmic}
    \label{algo:first-pass}
    
\end{algorithm}
% \vspace{-1ex}

The first pass, \emph{Declaration Collection}, constructs the symbol table for the target language (\cref{algo:first-pass}).
Each node is visited in preorder: declaration nodes (constants, variables, and relations) record their signatures in a shared context, while expression nodes (e.g., \texttt{And}, \texttt{Or}, \texttt{Not}, \texttt{Implies}) are traversed only to process their children without generating expressions.
Together, these recorded entries form a declaration environment, a mapping that registers all identifiers to their declared type or signature.
This environment ensures that every symbol referenced in the subsequent expression generation stage (e.g., variable names or predicate signatures) has been properly introduced.

\begin{algorithm}[t]
    \caption{Second pass to generate expression in an AST}
    \begin{algorithmic}
       \Function{GenerateExpression}{$r$}
            \If {$r$ = atomic sentence}
                \State $a \gets \emptyset$
                \For{$x \in r.args$ }
                   \State $a \gets a \cup \Call{GenerateExpression}{x}$
                \EndFor
                \State $name \gets r.name$
                \State \Return $\Call{AtomicSentence}{name,a}$
            \ElsIf {$r$ = binary sentence}
                \State $a \gets r.left\_operand$
                \State $a \gets \Call{GenerateExpression}{a}$
                \State $b \gets r.right\_operand$
                \State $b \gets \Call{GenerateExpression}{b}$
                \State $op \gets r.operator$
                \State \Return $\Call{BinarySentence}{a,op,b}$     
            \ElsIf {$r$ = negation sentence}
                \State $s \gets {r.sentence}$
                \State $n \gets \Call{GenerateExpression}{s}$
                \State \Return $\Call{NegatedSentence}{n}$
            \ElsIf {$r$ = quantified sentence}
                \State $q \gets r.quantifier$
                \State $s \gets \Call{GenerateExpression}{r.scope}$
                \State \Return $\Call{QuantifiedSentence}{q,s}$
            \ElsIf {$r$ = variable or constant}
                \State \Return $r.name$
            \EndIf
       \EndFunction
    \end{algorithmic}
    \label{algo:second-pass}
\end{algorithm}

The second pass, \emph{Expression Generation}, emits logical statements in the reasoning engine's target language (\cref{algo:second-pass}).
Each AST node is revisited to produce concrete code: constants and variables map to their declared identifiers, relational nodes expand into function calls with correct arity, and logical operators format operands according to the solver’s grammar.
Quantified nodes introduce scoped variables and generate expressions with explicit bindings, ensuring variable names remain unique and properly scoped.
All generated expressions are then added to the solver’s assertion set (e.g., \texttt{s.add(...)} in Z3).
This pass converts the FOLAST into final executable reasoning rules, grounding natural language in formal logic.

%% file: latex/sections/evaluation.tex
\subsection{Evaluation Setup}

\paragraph{Baseline}
We compare \tool against two representative baselines.
The first baseline is Grammar-Constrained Decoding (GCD)~\cite{gcd}, a state-of-the-art approach that enforces syntactic correctness by encoding the target formal language as a context-free grammar and constraining the LLM’s decoding process accordingly. We run GCD using the authors’ released implementation and follow the original evaluation protocol.

The second baseline is Logic-LM~\cite{logiclm}, a representative neuro-symbolic framework that translates natural language into first-order logic using few-shot prompting, without explicit syntactic or semantic constraints on the generated formulas. To evaluate whether \tool can strengthen existing neuro-symbolic systems, we replace Logic-LM’s original NL-to-FOL translation component with \tool while keeping all other components unchanged. For this comparison, we report both executable rate (i.e., whether the generated FOL formulas are syntactically valid and solver-executable) and downstream NLI accuracy.

For fairness, we evaluate using the same models (Gemma, Llama, Mistral, and Qwen) with sizes ranging from 0.5 to 27B parameters, as shown in \cref{tab:syntactic,tab:semantic,tab:logiclm_accuracy,tab:logiclm_executable}. For Logic-LM integration, we maintain the original pipeline but substitute only the FOL translation component.

\paragraph{Datasets}
We use three natural language inference (NLI) datasets, LogicNLI \cite{logicnli}, ProofWriter~\cite{proofwriter}, and FOLIO \cite{folio}.
These datasets are widely used for NLI \cite{logiclm, morishita2024enhancing} and first-order logic (FOL) translation tasks \cite{code4logic, yang-etal-2024-harnessing}.
In total, we use 3,000 premise-hypothesis pairs.
Each premise consists of a set of sentences, with each sentence corresponding to one logical rule.
The hypothesis is a single sentence expressible as one logical statement.
The sentence structures are relatively simple, with minimal co-reference and inter-sentence dependencies.
Each pair is labeled as \texttt{entailment}, \texttt{contradiction}, or \texttt{uncertain}.
To obtain the entailment prediction $\hat{y}$, \tool parses the premises $p$ and hypotheses $h$ to generate executable code for the automated reasoning engine \textsc{Solver}, as defined in \cref{eq:solver}.

\begin{equation}
\label{eq:solver}
\begin{array}{rl}
\textsc{Solver}(p,h)=&
\begin{cases}
\text{Ent.},& p \models h \wedge \ p \not\models \neg h, \\
\text{Cont.},& p \not\models h \wedge \ p \models \neg h, \\
\text{Unc.},& \text{otherwise}
\end{cases}
\\[1em]
\hat{y} = & \textsc{Solver}(p,h)
\end{array}
\end{equation}

\paragraph{RQs and Metrics}
Our evaluations aim to answer two research questions.

\begin{itemize}[leftmargin=*]
    \item \textbf{RQ1: Syntax correctness.}
    We assess whether \tool generates rules that adhere to first-order logic syntax, as specified in \cref{fig:grammar}.
    Syntax correctness is quantified using the correctness rate defined in \cref{eq:syntax}, where $N_{\text{correct}}$ is the number of sentences with correct syntax and $N_{\text{total}}$ is the total number of sentences.

    \begin{equation}
        \label{eq:syntax}
        \text{Syntax Correctness Rate} = \frac{N_{\text{correct}}}{N_{\text{total}}}
    \end{equation}

    \item \textbf{RQ2: Semantic correctness.}
    Following standard practice in prior NL-to-FOL work~\cite{gcd, code4logic}, we assess semantic correctness through downstream natural language inference (NLI) accuracy. This indirect evaluation is necessary because FOL expressions admit many truth-equivalent variants (e.g., $\neg (P \land Q) \equiv \neg P \lor \neg Q$), making direct comparison against a single canonical form impractical.
    Each NLI problem consists of premises and a hypothesis, and correctness is measured as \tool’s accuracy (\cref{eq:semantic}), defined as the proportion of predictions ($\hat{y_i}$) matching the gold labels ($y$).
    
    \begin{equation}
        \label{eq:semantic}
        \text{Accuracy} = \frac{\sum_{i=1}^{n}1(\hat{y_i} = y_i)}{n}
    \end{equation}

    \item \textbf{RQ3: Integration with neuro-symbolic systems.}
    We evaluate whether \tool’s AST-guided translation improves existing neuro-symbolic frameworks that rely on unconstrained zero-shot or few-shot prompting. Specifically, we integrate \tool into the Logic-LM \cite{logiclm} pipeline by replacing its original FOL translation module. Performance is assessed using the same two key metrics, which are syntactic and semantic accuracy.
    
\end{itemize}

\subsection{RQ1: Syntax Correctness}

\paragraph{Result} 

\cref{tab:syntactic} shows that \tool \textbf{consistently outperforms} the grammar-constrained decoding (GCD) baseline~\cite{gcd} across all model families and scales, achieving \textbf{near-perfect syntactic correctness} (up to 0.99).
The improvement is most pronounced in smaller models (e.g., Gemma-2-2B, Llama-3.2-1B), where syntax control is typically fragile. 
This demonstrates that \tool’s \emph{iterative parser} effectively mitigates syntax drift by incrementally constructing the abstract syntax tree (AST), thus reducing the model’s cognitive burden.
For larger models (e.g., Mistral-22B, Qwen-2.5-14B), \tool saturates near 0.99 accuracy, showing that once local node validity is guaranteed, \emph{global syntactic integrity emerges naturally}.

\begin{table}[h]
    \centering
    \small
    \begin{tabular}{|l|cc|cc|}
        \hline
         & \multicolumn{2}{|c|}{\textbf{FOLIO}} & \multicolumn{2}{|c|}{\textbf{LogicNLI}} \\
         \hline
         \textbf{Model} & \textbf{GCD} & \textbf{\makecell{\textsc{NL2}\\\textsc{Logic}}} & \textbf{GCD} & \textbf{\makecell{\textsc{NL2}\\\textsc{Logic}}} \\
         \hline
         gemma-2-2b & 0.56 & \textbf{0.92} & 0.25 & \textbf{0.94} \\
         gemma-2-9b & 0.77 & \textbf{0.98} & 0.91 & \textbf{0.99} \\
         gemma-2-27b & 0.81 & \textbf{0.99} & 0.93 & \textbf{0.99} \\
         \hline
         llama-3.2-1b & 0.27 & \textbf{0.99} & 0.08 & \textbf{0.97} \\
         llama-3.2-3b & 0.29 & \textbf{0.93} & 0.28 & \textbf{0.98} \\
         llama-3.1-8b & 0.13 & \textbf{0.96} & 0.68 & \textbf{0.98} \\
         \hline
         ministral-8b & 0.09 & \textbf{0.99} & 0.07 & \textbf{0.98} \\
         mistral-22b & 0.29 & \textbf{0.99} & 0.46 & \textbf{0.98} \\
         \hline
         qwen-2.5-0.5b & 0.14 & \textbf{0.66} & 0.05 & \textbf{0.80} \\
         qwen-2.5-1.5b & 0.36 & \textbf{0.87} & 0.09 & \textbf{0.99} \\
         qwen-2.5-3b & 0.12 & \textbf{0.81} & 0.01 & \textbf{0.99} \\
         qwen-2.5-7b & 0.01 & \textbf{0.94} & 0.01 & \textbf{0.99} \\
         qwen-2.5-14b & 0.53 & \textbf{0.92} & 0.56 & \textbf{0.99} \\
         \hline
    \end{tabular}
    \caption{Syntax correctness rate. \tool consistently outperforms GCD and achieves near-perfect syntax correctness.}
    \label{tab:syntactic}
\end{table}

\paragraph{Error Analysis}
However, errors may still occur when the LLM produces invalid nodes, rendering the accumulated AST incorrect.
Two types of errors are commonly observed in our experiments.
The first type violates the given JSON output format entirely, resulting in missing nodes, as illustrated in \cref{fig:invalid_schema}.
Although rare, the LLM may hallucinate and produce incomplete or unparseable JSON.
The second type conforms to the JSON schema but leaves required fields empty, resulting in invalid nodes, such as missing variable names in \texttt{Term} nodes or relation names in \texttt{Atom} nodes, as shown in \cref{fig:invalid_value}.
Together, these cases account for less than 5\% of all sentences.
Error statistics are summarized in \cref{tab:syntactic_error}.
We minimized these errors by employing strong output-control techniques, including the structured output feature in vLLM~\cite{vllm} and few-shot prompting (\cref{sec:system_prompts}) to guide the LLM in producing structured JSON representations of AST nodes.

\begin{figure}
    \centering
    \small
    \begin{subfigure}{\linewidth}
        \begin{mdframed}
            \ttfamily
            \{"quantifier": \{\\
            \hspace*{2em}"quantifier" : \{
            ...
        \end{mdframed}
        \vspace{-1ex}
        \caption{In rare cases, the LLM produces incomplete JSON that exceeds the maximum token limit.}
        \label{fig:invalid_schema}
    \end{subfigure}

    \vspace{1ex}
    
    \begin{subfigure}{\linewidth}
        \begin{mdframed}
            \ttfamily
            \{"transitive\_verb": "kind",\\
            \hspace*{1em}"subject": "sophia",\\
            \hspace*{1em}"object": ""
            \}
        \end{mdframed}
        \vspace{-1ex}
        \caption{The \texttt{parser} applies an incorrect JSON schema for the sentence “Sophia is kind”, causing the LLM to leave the object empty. As a result, the generated node is invalid.}
        \label{fig:invalid_value}
    \end{subfigure}

    \caption{Common errors that result in first-order logic syntax violations.}
    \label{fig:mistake_syntax}
\end{figure}

\begin{table}[h]
    \centering
    \small
    \begin{tabular}{|l|c|c|c|}
         \hline
         \textbf{Model} & \textbf{\makecell{Missing\\Nodes}} & \textbf{\makecell{Invalid\\Nodes}} & \textbf{\makecell{Total\\Sentences}} \\
         \hline
         gemma-2-2b & 15 & 472 & \multirow{13}{*}{7100} \\
         gemma-2-9b & 1 & 63 & \\
         gemma-2-27b & 1  & 24 &\\
         \cline{1-3}
         llama-3.2-1b & 127 & 5 & \\
         llama-3.2-3b & 133 & 148 & \\
         llama-3.1-8b & 107 & 97 & \\
         \cline{1-3}
         ministral-8b & 100 & 9 & \\
         mistral-22b & 100 & 13 & \\
         \cline{1-3}
         qwen-2.5-0.5b & 13 & 1786 & \\
         qwen-2.5-1.5b & 31 & 392 & \\
         qwen-2.5-3b & 18 & 530 & \\
         qwen-2.5-7b & 0 & 173 & \\
         qwen-2.5-14b & 5 & 224 & \\
         \hline
    \end{tabular}
    \caption{Error analysis on syntax correctness, accounting for less than 5\% of all sentences.}
    \label{tab:syntactic_error}
    \vspace{-1ex}
\end{table}

\subsection{RQ2: Semantic Correctness}

\paragraph{Result} 
\cref{tab:semantic} shows that \tool improves performance on natural language inference tasks by an average of 31\% over the grammar-constrained decoding (GCD) baseline.
This gain stems from \tool’s parser, which incrementally constructs the abstract syntax tree (AST) instead of generating the entire logical rule in one step.
The iterative design lowers the LLM’s cognitive load when identifying entities, predicates, and logical connectives, which is particularly advantageous for smaller models.
\tool achieves relatively higher accuracy on LogicNLI than on FOLIO because LogicNLI contains simpler sentence structures with clearer connectives and clauses, making the top-down parsing approach more natural.
In contrast, on FOLIO, some larger models with GCD surpass \tool, suggesting that excessive structural decomposition may not always be beneficial, especially for sentences that are not naturally parse-able in a top-down manner.

\begin{table}[h]
    \centering
    \small
    \begin{tabular}{|l|cc|cc|}
        \hline
         & \multicolumn{2}{|c|}{\textbf{FOLIO}} & \multicolumn{2}{|c|}{\textbf{LogicNLI}} \\
         \hline
         \textbf{Model} & \textbf{GCD} & \textbf{\makecell{\textsc{NL2}\\\textsc{Logic}}} & \textbf{GCD} & \textbf{\makecell{\textsc{NL2}\\\textsc{Logic}}} \\
         \hline
         gemma-2-2b & 0.27 & \textbf{0.35} & 0.22 & \textbf{0.41} \\
         gemma-2-9b & \textbf{0.53} & 0.38 & 0.23 & \textbf{0.37} \\
         gemma-2-27b & \textbf{0.61} & 0.40 & 0.24 & \textbf{0.37} \\
         \hline
         llama-3.2-1b & 0.17 & \textbf{0.38} & 0.11 & \textbf{0.33} \\
         llama-3.2-3b & 0.18 & \textbf{0.36} & 0.15 & \textbf{0.37} \\
         llama-3.1-8b & 0.12 & \textbf{0.37} & 0.17 & \textbf{0.37} \\
         \hline
         ministral-8b & 0.16 & \textbf{0.37} & 0.04 & \textbf{0.38} \\
         mistral-22b & 0.23 & \textbf{0.38} & 0.14 & \textbf{0.34} \\
         \hline
         qwen-2.5-0.5b & 0.17 & \textbf{0.26} & 0.12 & \textbf{0.36} \\
         qwen-2.5-1.5b & 0.19 & \textbf{0.34} & 0.03 & \textbf{0.41} \\
         qwen-2.5-3b & 0.08 & \textbf{0.32} & 0.01 & \textbf{0.35} \\
         qwen-2.5-7b & 0.01 & \textbf{0.37} & 0.01 & \textbf{0.34} \\
         qwen-2.5-14b & 0.34 & \textbf{0.37} & 0.27 & \textbf{0.35} \\
         \hline
    \end{tabular}
    \caption{Semantic correctness measured by accuracy on natural language inference (NLI) tasks.}
    \label{tab:semantic}
\end{table}

\begin{figure}[h]
    \centering
    \small
    \begin{subfigure}{\linewidth}
        \begin{mdframed}
            \ttfamily
            Parsing 'John is unable to walk'\\
            Unary operator parser. Operator: 'Not'\\
            |--- Parsing 'John is not unable to walk'\\
            \hspace*{1em} Unary operator parser. Operator: 'Not'\\
            \hspace*{3.5em}... 
        \end{mdframed}
        \vspace{-1ex}
        \caption{Infinite recursion caused by parsing words with implicit negation.}
        \label{fig:infinite_recursion}
    \end{subfigure}

    \vspace{1ex}

    \begin{subfigure}{\linewidth}
    \centering
        \includegraphics[]{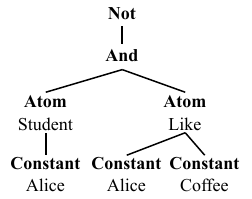}
        \caption{Incorrect operator ordering when parsing sentences with complex compound sentences. The sentence is \texttt{"Alice is not a student and does not like coffee"}, where \texttt{And} operator should precede the \texttt{Not } operator.}
        \label{fig:incorrect_order}
    \end{subfigure}
    \caption{Common semantic errors.}
    \vspace{-2ex}
\end{figure}

\paragraph{Observation and Error Analysis}
There are two common semantic errors.
First, words with implicit negation (e.g., unable, unaware, inconsistent) often confuse the parser.
Despite careful prompt design, the parser sometimes misinterprets them and enters an infinite recursive parse.
For instance, the sentence \texttt{'John is unable to walk'} is parsed back and forth to the sentence \texttt{'John is not unable to walk'}, as illustrated in \cref{fig:infinite_recursion}.
Second, sentences with subtle operator ordering often lead to operator order errors.
For example, the sentence Alice is not a student and does not like coffee is occasionally parsed with the \texttt{Not} operator preceding \texttt{And}, producing a semantically incorrect AST, as shown in \cref{fig:incorrect_order}.

\subsection{RQ3: Integration with Neuro-Symbolic Systems}

To demonstrate \tool's practical value as a modular component, we integrated it into Logic-LM~\cite{logiclm}, a representative neuro-symbolic framework combining LLMs with symbolic solvers. Logic-LM originally uses few-shot prompting for FOL translation without explicit syntactic or semantic constraints. We replaced this module with \tool while keeping all other pipeline components unchanged (Z3 solver, refinement mechanisms, answer selection), isolating the impact of our AST-guided translation on syntactic validity and downstream reasoning accuracy.

Table~\ref{tab:logiclm_executable} presents executable rate comparison on ProofWriter~\cite{proofwriter} and FOLIO~\cite{folio}. Logic-LM's original few-shot translation produces highly variable executable rates (0.01-0.94), with particularly poor performance on smaller models (e.g., 0.01 for gemma-2-2b, qwen-2.5-0.5b on ProofWriter). \tool achieves near-perfect executable rate (0.99) across all 13 models and both datasets through iterative AST construction validating each logical construct incrementally.

This syntactic validity improvement translates into better semantic correctness. Table~\ref{tab:logiclm_accuracy} shows accuracy on executable rules only, where predictions originate from logic solver execution rather than backup strategies (random guessing or CoT fallback). \tool improves Logic-LM's accuracy by $\sim$31\% on average. Gains are most pronounced on smaller models: on ProofWriter, gemma-2-2b improves from 0.01 to 0.58, llama-3.2-3b from 0.03 to 0.68, qwen-2.5-7b from 0.08 to 0.90, demonstrating that \tool's structured decomposition reduces cognitive load by parsing natural language into FOL one clause at a time.

\begin{table}[h]
    \centering
    \small
    \begin{tabular}{|l|cc|cc|}
        \hline
         & \multicolumn{2}{|c|}{\textbf{ProofWriter}} & \multicolumn{2}{|c|}{\textbf{FOLIO}} \\
         \hline
         \textbf{Model} & \textbf{\makecell{Logic\\LM}} & \textbf{\makecell{NL2\\Logic}} & \textbf{\makecell{Logic\\LM}} & \textbf{\makecell{NL2\\Logic}} \\
         \hline
         gemma-2-2b & 0.01 & \textbf{0.99} & 0.07 & \textbf{0.99} \\
         gemma-2-9b & 0.90 & \textbf{0.99} & 0.65 & \textbf{0.99} \\
         gemma-2-27b & 0.94 & \textbf{0.99} & 0.60 & \textbf{0.99} \\
         \hline
         llama-3.2-1b & 0.72 & \textbf{0.99} & 0.83 & \textbf{0.99} \\
         llama-3.2-3b & 0.25 & \textbf{0.99} & 0.38 & \textbf{0.99} \\
         llama-3.1-8b & 0.90 & \textbf{0.99} & 0.58 & \textbf{0.99} \\
         \hline
         ministral-8b & 0.20 & \textbf{0.99} & 0.14 & \textbf{0.99} \\
         mistral-22b & 0.85 & \textbf{0.99} & 0.65 & \textbf{0.99} \\
         \hline
         qwen-2.5-0.5b & 0.01 & \textbf{0.99} & 0.01 & \textbf{0.99} \\
         qwen-2.5-1.5b & 0.01 & \textbf{0.99} & 0.05 & \textbf{0.99} \\
         qwen-2.5-3b & 0.45 & \textbf{0.99} & 0.52 & \textbf{0.99} \\
         qwen-2.5-7b & 0.15 & \textbf{0.99} & 0.55 & \textbf{0.99} \\
         qwen-2.5-14b & 0.38 & \textbf{0.99} & 0.72 & \textbf{0.99} \\
         \hline
    \end{tabular}
    \caption{Executable rate: Logic-LM original vs. with \tool integration.}
    \label{tab:logiclm_executable}
\end{table}

\begin{table}[h]
    \centering
    \small
    \begin{tabular}{|l|cc|cc|}
        \hline
         & \multicolumn{2}{|c|}{\textbf{ProofWriter}} & \multicolumn{2}{|c|}{\textbf{FOLIO}} \\
         \hline
         \textbf{Model} & \textbf{\makecell{Logic\\LM}} & \textbf{\makecell{NL2\\Logic}} & \textbf{\makecell{Logic\\LM}} & \textbf{\makecell{NL2\\Logic}} \\
         \hline
         gemma-2-2b & 0.01 & \textbf{0.58} & 0.21 & \textbf{0.36} \\
         gemma-2-9b & 0.45 & \textbf{0.85} & 0.26 & \textbf{0.39} \\
         gemma-2-27b & 0.60 & \textbf{0.90} & 0.31 & \textbf{0.41} \\
         \hline
         llama-3.2-1b & 0.25 & \textbf{0.47} & 0.18 & \textbf{0.38} \\
         llama-3.2-3b & 0.03 & \textbf{0.68} & 0.14 & \textbf{0.36} \\
         llama-3.1-8b & 0.40 & \textbf{0.91} & 0.22 & \textbf{0.38} \\
         \hline
         ministral-8b & 0.18 & \textbf{0.75} & 0.04 & \textbf{0.39} \\
         mistral-22b & 0.51 & \textbf{0.78} & 0.26 & \textbf{0.41} \\
         \hline
         qwen-2.5-0.5b & 0.01 & \textbf{0.38} & 0.01 & \textbf{0.26} \\
         qwen-2.5-1.5b & 0.01 & \textbf{0.48} & 0.02 & \textbf{0.34} \\
         qwen-2.5-3b & 0.17 & \textbf{0.37} & 0.14 & \textbf{0.32} \\
         qwen-2.5-7b & 0.08 & \textbf{0.90} & 0.22 & \textbf{0.37} \\
         qwen-2.5-14b & 0.32 & \textbf{0.86} & 0.31 & \textbf{0.38} \\
         \hline
    \end{tabular}
    \caption{Accuracy on executable rules: Logic-LM original vs. with \tool integration.}
    \label{tab:logiclm_accuracy}
\end{table}

%% file: latex/sections/conclusion.tex
This paper presents \tool, an AST-guided framework for translating natural language into first-order logic using large language models (LLMs).
Unlike prior work that treats this task as free-form text generation with limited control, \tool incrementally constructs the AST in a top-down manner.
This approach provides stronger control over the LLM output, achieving near-perfect syntactic correctness, improving semantic accuracy by 30\% on LogicNLI, FOLIO, and ProofWriter datasets, and improving existing neuro-symbolic system Logic-LM by 31\% on both syntactic and downstream reasoning task accuracy.

%% file: latex/sections/ethics.tex
\paragraph{Datasets and Models} 
We rely solely on publicly available datasets and models, without involving human subjects or newly collected data. 
No personal, private, or sensitive information is used. 
Therefore, this work poses no risks related to privacy, consent, or data annotation ethics. 
All datasets and models are utilized in accordance with their respective licenses.

\paragraph{Potential Misleading Proof}
We envision that rules generated by \tool will be executed using automated reasoning engines such as Z3, PyProver, or SMT-LIB to verify a hypothesis against a set of premises.  
Although the goal is to obtain a formally provable answer, the proof must be interpreted carefully.  
Given a set of premises $P$ and a hypothesis $Q$, $P \models Q$ does not necessarily imply that $Q$ is proven.  
If $P \models \neg Q$ also holds, the result is \textit{uncertain} instead of \textit{entailment}.  
If the premises themselves are unsatisfiable (e.g., due to semantic errors during translation), $P$ may entail any statement.  
Hence, one must examine all possible outcomes, which are $P \models Q$, $P \models \neg Q$, and even the satisfiability of $P$ itself.

\paragraph{Use of AI Assistants}
We acknowledge the use of AI assistants for grammar checking.
The authors remain fully responsible for the scientific contributions, experimental results, and all claims presented in this paper.

%% file: latex/sections/limitations.tex
\paragraph{Dependencies on Multiple Sentences}
\tool converts text into first-order logic, one sentence at a time. 
It ensures consistent predicate arity, constant' names, and variable' names across sentences by prompting the LLM to use the base form of each word, such as the present tense for verbs and objects without modifiers. 
However, certain sentences require contextual information from adjacent sentences or the entire text to generate accurate logical representations. 
In such cases, \tool does not yet handle co-reference or implicit relational links across sentences. 
For instance, given the premise “John is the father of Alice” and the hypothesis “John is the parent of Alice,” separate translation would yield two distinct relations: \texttt{father} and \texttt{parent}.
With full context, however, the predicate \texttt{father} could instead be expressed as a conjunction of \texttt{parent} and \texttt{male}.

\paragraph{Reliance on Structured Generations}
\tool relies on structured output capabilities available in both commercial APIs (e.g., OpenAI) and open-source LLM serving frameworks (e.g., vLLM). 
It prompts the LLM to generate JSON outputs following schemas specific to each parser type. 
However, not all LLM serving frameworks support this feature natively.
For instance, \texttt{llama.cpp} requires integration with the \texttt{lm-format-enforcer} tool to enable structured output.

%% file: latex/sections/acknowledgement.tex
We would like to thank the anonymous reviewers for their constructive comments and suggestions.
This work is supported in part by the National Science Foundation under grant CNS-2442171. Any opinions, findings, and conclusions
made in this paper are those of the authors and do not
necessarily reflect the views of the funding agencies.

%% file: latex/sections/appendix.tex
\tcbset{
  colback=white,
  colframe=black!20,
  boxrule=0.5pt,
  arc=2pt,
  outer arc=2pt,
  left=4pt, right=4pt, top=3pt, bottom=3pt,
  fontupper=\ttfamily\small,
  enhanced jigsaw,
  breakable
}

The \texttt{parser}, described in \cref{sec:design} and illustrated in \cref{fig:overview}, iteratively constructs the abstract syntax tree (AST) by recursively analyzing each sentence component through specialized submodules: \texttt{ParserSelector}, \texttt{AtomicSentenceParser}, \texttt{QuantifiedSentenceParser}, and \texttt{LogicalSen\linebreak tenceParser}.
The \texttt{ParserSelector} determines the appropriate parser for a sentence, while the remaining three parsers perform parsing and recursively invoke \texttt{the ParserSelector} for any child nodes.
The system prompt for \texttt{ParserSelector} is shown in 
\cref{fig:sp_parserselector}. 
While the  system prompt for \texttt{the QuantifiedSentenceParser} is shown in \cref{fig:sp_quantified}.
The \texttt{LogicalSentenceParser} involves multiple system prompts, as shown in \cref{fig:sp_logical_binary,fig:sp_logical_unary}. 
\texttt{AtomicSentenceParser} also involves multiple system prompts, as shown in \cref{fig:sp_atom1,fig:sp_atom2,fig:sp_atom3,fig:sp_atom4,fig:sp_atom5}.

\begin{figure}[h]
    % Choose Parser System Prompt
    \begin{tcolorbox}[title={\texttt{ParserSelector} System Prompt}]
    You are an expert on classifying the sentence by its overall structure:\\ 
    A = An atomic logical statement. (no quantifiers, no logical connectives, no negation)\\ 
    B = A quantified logical statement when the sentence talks about a general rule that covers many entities.\\ 
    If the sentence only mentions specific proper names, or an instances of variable (x,y,z), or if quantifiers appear only inside part of the sentence, then it should be classified as A, C, or D instead.\\ 
    C = A compound logical sentence, where each part is connected with logical connectives such as 'and', 'or', 'if...then', or 'only if'.\\ 
    D = A statement that contains literal negation of another sentence ('not', 'no', 'dont', 'doesnt'). Only look for a literal negation.
    \\\\ 
    Example 1:\\ 
    Sentence: 'Alice sings.'\\ 
    Answer: \{ "answer" : "A"\} 
    \\
    ...
    \\
    Now, it is your turn
    \\\\
    Input: \{input sentence\}\\
    Answer: 
    \end{tcolorbox}

    \caption{The system prompt for selecting the appropriate parser. The first step in parsing a sentence is to classify whether it is an atomic, quantified, compound, or negation logical sentence.}
    \label{fig:sp_parserselector}
\end{figure}

% Quantified System Prompt
\begin{figure}[h]
    \begin{tcolorbox}[title={\texttt{QuantifiedSentenceParser} System Prompt}]
    You identified the sentence as a quantified logical statement.
    \\\\
    Task:\\
    1. Select the correct quantifier:\\
       - ForAll (e.g., all, every, each, no one) -> the logical statement applies to ALL entities\\
       - ThereExists (e.g., some, there is, at least one, a) -> the logical statement applies to SOME entities\\
    2. Identify the variable (and all reference) being quantified (the noun phrase that follows the quantifier, e.g., "student", "person", "dog") and replace it with a letter like x, y, or z.\\
    3. Rewrite the sentence WITHOUT the quantifier, keeping the variable in place so the sentence is still natural and understandable. Preserve the exact wording and capitalization of all subject and object names. If there is multiple quantifier, just remove the outermost.\\
    4. If the sentence is ambiguous, you should rephrase it so that the next parser will understand whether it is an atomic logical sentence, or logical sentence with connectives, or a quantified logical sentence.
    \\\\
    Examples:
    \\\\
    Input: "All students study hard."\\
    Output: 
    \{ "quantifier" : "ForAll", "variable" : "x", "sentence\_without\_quantifier" : "x study hard."\}
    \\\\
    ...
    \\\\
    Now, it is your turn
    \\\\
    Input: \{input sentence\}\\
    Output: 
    \end{tcolorbox}
    \caption{The system prompt for parsing a quantified logical sentence.
    It instructs the LLM to extract the quantifier, variable, and the scoped logical sentence.
    }
    \label{fig:sp_quantified}
\end{figure}

\begin{figure}[h]
    % Binary Logical System Prompt
    \begin{tcolorbox}[title={\texttt{LogicalSentenceParser} System Prompt}]
    You parse a sentence into its OUTERMOST (top-level) logical operator and its two operands. 
    Always choose the operator that governs the entire sentence (outermost scope). Do not parse nested or inner operators here.
    Each operand, left and right, are a standalone and complete sentence, not just a phrase, meaning it has at least subject and verb/to be.
    Output JSON matching:\\ 
    operator: one of ["Not","And","Or","If",
    "OnlyIf","IfAndOnlyIf"]\\ 
    left\_operand: rewrite the left part as a clean, standalone clause. Preserve the exact wording and capitalization of all subject and object names. But, resolve co-reference such as she, he, it, etc.\\ 
    right\_operand: rewrite the right part as a clean, standalone clause. Preserve the exact wording and capitalization of all subject and object names. But, resolve co-reference such as she, he, it, etc.\\ 
    \\ 
    Decide by the main structure of the whole sentence:\\ 
    - And: two clauses joined by and.\\ 
    - Or: two clauses joined by or / either ... or.\\ 
    - If: conditional “if … then …” (antecedent = left, consequent = right).\\ 
    - OnlyIf: “P only if Q” (left = P, right = Q; Q is required for P).\\ 
    - IfAndOnlyIf: “iff / if and only if / exactly when / just in case”.\\ 
    \\ 
    Examples: 
    \\\\ 
    Input: "Alice sings and dances."\
    Output: \{"operator": "And", "left\_operand": "Alice sings", "right\_operand": "Alice dances"\} 
    \\\\ 
    ...
    \\
    Now, it is your turn
    \\\\
    Input: \{input sentence\}\\
    Output: 
    \end{tcolorbox}
    \caption{The system prompt for parsing a sentence involving a binary logical operator.
    It instructs the LLM to extract the operator and its two operands.}
    \label{fig:sp_logical_binary}
\end{figure}

\begin{figure}[h]
    % Unary Logical System Prompt
    \begin{tcolorbox}[title={\texttt{LogicalSentenceParser} System Prompt}]
    You parse a sentence whose top-level operator is unary negation.
    \\\\
    Output JSON matching:\\
    operator: always "Not"\\
    operand : rewrite the sentence without the outermost negation
    \\\\
    Guidelines:\\
    - Always set operator = "Not".\\
    - For the operand, remove only the **outermost** negation. \\
    - If there are multiple negations, strip just the outermost one and keep the inner ones.\\
    - Rewrite the operand as a natural, grammatical sentence.\\
    - Do not add explanations or extra words.\\
    - Return JSON only.
    \\\\
    Examples:
    \\\\
    Input: "It is not raining."\\
    Output: \{"operator": "Not", "operand": "It is raining"\}
    \\\\
    Input: "No student is absent."\\
    Output: \{"operator": "Not", "operand": "a student is absent"\}
    \\\\
    Input: "It is not true that John is not guilty."
    Output: \{"operator": "Not", "operand": "John is not guilty"\}
    \\\\
    Input: "Nobody loves me."
    Output: \{"operator": "Not", "operand": "Somebody loves me"\}
    \\\\
    Now, it is your turn
    \\\\
    Input: \{input sentence\}\\
    Answer: 
    \end{tcolorbox}
    \caption{The system prompt for parsing a sentence involving a negation operator.
It instructs the LLM to extract the sentence excluding the negation.}
    \label{fig:sp_logical_unary}
\end{figure}

\begin{figure}[h]
    % Choose Relation System Prompt
    \begin{tcolorbox}[title={\texttt{AtomicSentenceParser} System Prompt}]
    You classify an ATOMIC natural-language predicate into one of:\\
    A = Adjective/property\\
    B = Intransitive verb (takes no object)\\
    C = Transitive verb (takes exactly one object)\\
    D = Ditransitive verb (takes two objects)\\
    \\
    Examples:\\
    \\
    Input: "Alice is tall."\\
    Output: \{"answer":"A"\}\\
    \\
    Input: "Alice is a student."\\
    Output: \{"answer":"A"\}\\
    \\
    Input: "Alice runs."\\
    Output: \{"answer":"B"\}\\
    \\
    Input: "The baby sleeps."\\
    Output: \{"answer":"B"\}\\
    \\
    \\
    ...
    \\
    Now, it is your turn
    \\\\
    Input: \{input sentence\}\\
    Output: 
    \end{tcolorbox}
    \caption{The system prompt for parsing an atomic logical sentence.
It first instructs the LLM to determine whether the sentence involves an adjective, intransitive verb, transitive verb, or ditransitive verb predicate.}
    \label{fig:sp_atom1}
\end{figure}

\begin{figure}[h]
    % Adjective System Prompt
    \begin{tcolorbox}[title={\texttt{AtomicSentenceParser} System Prompt}]
    You extract the object and its adjective property from a simple atomic sentence.
    \\\
    Output JSON matching:\\
    adjective: the describing word or phrase. use the base form with no modifier\\
    obj: the entity being described (keep wording and capitalization verbatim). use the base form with no modifier\\
    \\
    Rules:\\
    - Handle only atomic adjective/property sentences (e.g., "Alice is tall", "The dog is happy").\\
    - Subjects must be individual names or noun phrases without quantifiers ("all", "every", "some", "no").\\
    - Adjective may be single word or short phrase (e.g., "absent", "very tall").\\
    - Do not paraphrase or change casing; copy terms exactly.\\
    - Ignore tense/negation; just extract subject and adjective.\\
    \\
    Examples:\\
    \\
    Input: "Alice is tall."\\
    Output: \{"adjective": "tall", "obj": "Alice"\}\\
    \\
    Input: "student is awesome."\\
    Output: \{"adjective": "awesome", "obj": "student"\}\\
    \\
    Input: "Bob is very tired."\\
    Output: \{"adjective": "very tired", "obj": "Bob"\}\\
    \\\\
    Now, it is your turn
    \\\\
    Input: \{input sentence\}\\
    Answer: 
    \end{tcolorbox}
    \caption{The system prompt for parsing an atomic logical sentence involving a subject and an adjective.
It instructs the LLM to extract the subject and the adjective predicate.}
    \label{fig:sp_atom2}
\end{figure}

\begin{figure}[h]
    % Intransitive System Prompt
    \begin{tcolorbox}[title={\texttt{AtomicSentenceParser} System Prompt}]
    You extract the subject and the main intransitive verb from a simple atomic sentence.\\
    \\
    Output JSON matching:\\
    verb: the main intransitive verb (copy exactly as in the input). use the verb base form\\
    subject: the entity performing the action (keep wording and capitalization verbatim)\\
    \\
    Rules:\\
    - Handle only atomic intransitive sentences (a subject + intransitive verb, with no object, no quantifier, no negation).\\
    - Subject is a proper name or noun phrase (e.g., "Alice", "The student").\\
    - Verb must appear exactly as written in the sentence (respect tense/aspect: "runs", "is running", "slept").\\
    - Do not paraphrase or alter capitalization.\\
    - Sentences with objects, quantifiers, or negation are out of scope.\\
    \\
    Examples:\\
    \\
    Input: "Alice runs."\\
    Output: \{"verb": "run", "subject": "Alice"\}\\
    \\
    Input: "The student sleeps."\\
    Output: \{"verb": "sleep", "subject": "The student"\}\\
    \\
    Input: "Bob is running."\\
    Output: \{"verb": "run", "subject": "Bob"\}\\
    \\
    Input: "Alice swam."\\
    Output: \{"verb": "swim", "subject": "Alice"\}\\
    \\
    Now, it is your turn
    \\\\
    Input: \{input sentence\}\\
    Answer: 
    \end{tcolorbox}
    \caption{The system prompt for parsing an atomic logical sentence involving a simple subject and intransitive verb structure.
It instructs the LLM to extract the subject and the verb predicate.}
    \label{fig:sp_atom3}
\end{figure}

\begin{figure}[h]
    % Transitive System Prompt
    \begin{tcolorbox}[title={\texttt{AtomicSentenceParser} System Prompt}]
    You extract the subject, the main transitive verb, and its single object from a simple atomic sentence.\\
    \\
    Output JSON matching:\\
    subject: the entity performing the action (copy wording and capitalization verbatim)\\
    verb: the main transitive verb. use the verb base form\\
    obj: the object of the verb (copy wording and capitalization verbatim). use the base form in infinitive form\\
    \\
    \\
    Examples:\\
    \\
    Input: "Alice loves Bob."\\
    Output: \{"subject": "Alice", "verb": "love", "obj": "Bob"\}\\
    \\
    Input: "The student reads a book."\\
    Output: \{"subject": "The student", "verb": "read", "obj": "a book"\}\\
    \\
    Input: "Bob is watching TV."\\
    Output: \{"subject": "Bob", "verb": "watch", "obj": "TV"\}\\
    \\
    Input: "Mary wrote a letter."\\
    Output: \{"subject": "Mary", "verb": "write", "obj": "a letter"\}\\
    \\
    Input: "John loves swimming."\\
    Output: \{"subject": "John", "verb": "love", "obj": "to swim"\}\\
    \\
    Input: "Doe likes to read a book"\\
    Output: \{"subject": "Doe", "verb": "like", "obj": "to read a book"\}\\
    \\
    Now, it is your turn
    \\\\
    Input: \{input sentence\}\\
    Answer: 
    \end{tcolorbox}
    \caption{The system prompt for parsing an atomic logical sentence involving a subject and a transitive verb.
It instructs the LLM to extract the subject, the verb predicate, and the object.}
    \label{fig:sp_atom4}
\end{figure}

\begin{figure}[h]
    % Ditransitive System Prompt
    \begin{tcolorbox}[title={\texttt{AtomicSentenceParser} System Prompt}]
    You extract the subject, the main ditransitive verb, its indirect object, and its direct object from a simple atomic sentence.\\
    \\
    Output JSON matching:\\
    subject: the entity performing the action (copy wording and capitalization verbatim)\\
    verb: the main ditransitive verb. use the base verb form\\
    indirect\_obj: the recipient/beneficiary of the action (copy wording and capitalization verbatim). use the infinitive form if needed\\
    direct\_obj: the thing being given/sent/shown/etc. (copy wording and capitalization verbatim). use the infinitive form if needed\\
    \\
    Examples:\\
    \\
    Input: "John gave Mary a book."\\
    Output: \{"subject": "John", "verb": "give", "indirect\_obj": "Mary", "direct\_obj": "a book"\}\\
    \\
    Input: "Alice sent Bob a letter."\\
    Output: \{"subject": "Alice", "verb": "send", "indirect\_obj": "Bob", "direct\_obj": "a letter"\}
    
    Input: "The teacher showed the students a picture."
    Output: \{"subject": "The teacher", "verb": "show", "indirect\_obj": "the students", "direct\_obj": "a picture"\}\\
    \\
    Input: "John gave a book to Mary."\\
    Output: \{"subject": "John", "verb": "give", "indirect\_obj": "Mary", "direct\_obj": "a book"\}\\
    \\
    ...
    \\
    Now, it is your turn
    \\\\
    Input: \{input sentence\}\\
    Answer: 
    \end{tcolorbox}
    \caption{The system prompt for parsing an atomic logical sentence involving a subject and a ditransitive verb.
It instructs the LLM to extract the subject, the verb predicate, the direct object, and the indirect object.}
    \label{fig:sp_atom5}
\end{figure}

%% file: custom.bib
@inproceedings{code4logic,
    title = "Few-Shot Natural Language to First-Order Logic Translation via Code Generation",
    author = "Liu, Junnan",
    editor = "Chiruzzo, Luis  and
      Ritter, Alan  and
      Wang, Lu",
    booktitle = "Proceedings of the 2025 Conference of the Nations of the Americas Chapter of the Association for Computational Linguistics: Human Language Technologies (Volume 1: Long Papers)",
    month = apr,
    year = "2025",
    address = "Albuquerque, New Mexico",
    publisher = "Association for Computational Linguistics",
    url = "https://aclanthology.org/2025.naacl-long.547/",
    doi = "10.18653/v1/2025.naacl-long.547",
    pages = "10939--10960",
    ISBN = "979-8-89176-189-6"
}

@inproceedings{folio,
  title={FOLIO: Natural Language Reasoning with First-Order Logic},
  author={Han, Simeng and Schoelkopf, Hailey and Zhao, Yilun and Qi, Zhenting and Riddell, Martin and Zhou, Wenfei and Coady, James and Peng, David and Qiao, Yujie and Benson, Luke and others},
  booktitle={Proceedings of the 2024 Conference on Empirical Methods in Natural Language Processing},
  pages={22017--22031},
  year={2024}
}

@inproceedings{logicnli,
  title={Diagnosing the first-order logical reasoning ability through LogicNLI},
  author={Tian, Jidong and Li, Yitian and Chen, Wenqing and Xiao, Liqiang and He, Hao and Jin, Yaohui},
  booktitle={Proceedings of the 2021 Conference on Empirical Methods in Natural Language Processing},
  pages={3738--3747},
  year={2021}
}

@inproceedings{gcd,
  title={Grammar-constrained decoding makes large language models better logical parsers},
  author={Raspanti, Federico and Ozcelebi, Tanir and Holenderski, Mike},
  booktitle={Proceedings of the 63rd Annual Meeting of the Association for Computational Linguistics (Volume 6: Industry Track)},
  pages={485--499},
  year={2025}
}

@inproceedings{sat,
    title = "Segment Any Text: A Universal Approach for Robust, Efficient and Adaptable Sentence Segmentation",
    author = "Frohmann, Markus  and
      Sterner, Igor  and
      Vuli{\'c}, Ivan  and
      Minixhofer, Benjamin  and
      Schedl, Markus",
    editor = "Al-Onaizan, Yaser  and
      Bansal, Mohit  and
      Chen, Yun-Nung",
    booktitle = "Proceedings of the 2024 Conference on Empirical Methods in Natural Language Processing",
    month = nov,
    year = "2024",
    address = "Miami, Florida, USA",
    publisher = "Association for Computational Linguistics",
    url = "https://aclanthology.org/2024.emnlp-main.665",
    pages = "11908--11941"
}

@inproceedings{wtpsplit,
    title = "Where{'}s the Point? Self-Supervised Multilingual Punctuation-Agnostic Sentence Segmentation",
    author = "Minixhofer, Benjamin  and
      Pfeiffer, Jonas  and
      Vuli{\'c}, Ivan",
    booktitle = "Proceedings of the 61st Annual Meeting of the Association for Computational Linguistics (Volume 1: Long Papers)",
    month = jul,
    year = "2023",
    address = "Toronto, Canada",
    publisher = "Association for Computational Linguistics",
    url = "https://aclanthology.org/2023.acl-long.398",
    pages = "7215--7235"
}

@inproceedings{yang-etal-2024-harnessing,
    title = "Harnessing the Power of Large Language Models for Natural Language to First-Order Logic Translation",
    author = "Yang, Yuan  and
      Xiong, Siheng  and
      Payani, Ali  and
      Shareghi, Ehsan  and
      Fekri, Faramarz",
    editor = "Ku, Lun-Wei  and
      Martins, Andre  and
      Srikumar, Vivek",
    booktitle = "Proceedings of the 62nd Annual Meeting of the Association for Computational Linguistics (Volume 1: Long Papers)",
    month = aug,
    year = "2024",
    address = "Bangkok, Thailand",
    publisher = "Association for Computational Linguistics",
    url = "https://aclanthology.org/2024.acl-long.375/",
    doi = "10.18653/v1/2024.acl-long.375",
    pages = "6942--6959"
}

@inproceedings{logiclm,
    title = "Logic-{LM}: Empowering Large Language Models with Symbolic Solvers for Faithful Logical Reasoning",
    author = "Pan, Liangming  and
      Albalak, Alon  and
      Wang, Xinyi  and
      Wang, William",
    editor = "Bouamor, Houda  and
      Pino, Juan  and
      Bali, Kalika",
    booktitle = "Findings of the Association for Computational Linguistics: EMNLP 2023",
    month = dec,
    year = "2023",
    address = "Singapore",
    publisher = "Association for Computational Linguistics",
    url = "https://aclanthology.org/2023.findings-emnlp.248/",
    doi = "10.18653/v1/2023.findings-emnlp.248",
    pages = "3806--3824"
}

@article{morishita2024enhancing,
  title={Enhancing reasoning capabilities of llms via principled synthetic logic corpus},
  author={Morishita, Terufumi and Morio, Gaku and Yamaguchi, Atsuki and Sogawa, Yasuhiro},
  journal={Advances in Neural Information Processing Systems},
  volume={37},
  pages={73572--73604},
  year={2024}
}

@inproceedings{vllm,
  title={Efficient Memory Management for Large Language Model Serving with PagedAttention},
  author={Woosuk Kwon and Zhuohan Li and Siyuan Zhuang and Ying Sheng and Lianmin Zheng and Cody Hao Yu and Joseph E. Gonzalez and Hao Zhang and Ion Stoica},
  booktitle={Proceedings of the ACM SIGOPS 29th Symposium on Operating Systems Principles},
  year={2023}
}

@inproceedings{bos2005recognising,
  title={Recognising textual entailment with logical inference},
  author={Bos, Johan and Markert, Katja},
  booktitle={Proceedings of Human Language Technology Conference and Conference on Empirical Methods in Natural Language Processing},
  pages={628--635},
  year={2005}
}

@article{zettlemoyer2012learning,
  title={Learning to map sentences to logical form: Structured classification with probabilistic categorial grammars},
  author={Zettlemoyer, Luke S and Collins, Michael},
  journal={arXiv preprint arXiv:1207.1420},
  year={2012}
}

@article{barker2009dimensions,
  title={Dimensions of Difficulty in Translating Natural Language into First Order Logic.},
  author={Barker-Plummer, Dave and Cox, Richard and Dale, Robert},
  journal={International Working Group on Educational Data Mining},
  year={2009},
  publisher={ERIC}
}

@article{abzianidze2017langpro,
  title={LangPro: Natural language theorem prover},
  author={Abzianidze, Lasha},
  journal={arXiv preprint arXiv:1708.09417},
  year={2017}
}

@inproceedings{lu2022parsing,
  title={Parsing natural language into propositional and first-order logic with dual reinforcement learning},
  author={Lu, Xuantao and Liu, Jingping and Gu, Zhouhong and Tong, Hanwen and Xie, Chenhao and Huang, Junyang and Xiao, Yanghua and Wang, Wenguang},
  booktitle={Proceedings of the 29th International Conference on Computational Linguistics},
  pages={5419--5431},
  year={2022}
}

@article{cao2019semantic,
  title={Semantic parsing with dual learning},
  author={Cao, Ruisheng and Zhu, Su and Liu, Chen and Li, Jieyu and Yu, Kai},
  journal={arXiv preprint arXiv:1907.05343},
  year={2019}
}

@article{wei2022chain,
  title={Chain-of-thought prompting elicits reasoning in large language models},
  author={Wei, Jason and Wang, Xuezhi and Schuurmans, Dale and Bosma, Maarten and Xia, Fei and Chi, Ed and Le, Quoc V and Zhou, Denny and others},
  journal={Advances in neural information processing systems},
  volume={35},
  pages={24824--24837},
  year={2022}
}

@article{kojima2022large,
  title={Large language models are zero-shot reasoners},
  author={Kojima, Takeshi and Gu, Shixiang Shane and Reid, Machel and Matsuo, Yutaka and Iwasawa, Yusuke},
  journal={Advances in neural information processing systems},
  volume={35},
  pages={22199--22213},
  year={2022}
}

@misc{creswell2205selection,
      title={Selection-Inference: Exploiting Large Language Models for Interpretable Logical Reasoning}, 
      author={Antonia Creswell and Murray Shanahan and Irina Higgins},
      year={2022},
      eprint={2205.09712},
      archivePrefix={arXiv},
      primaryClass={cs.AI},
      url={https://arxiv.org/abs/2205.09712}, 
}

@article{wang2023grammar,
  title={Grammar prompting for domain-specific language generation with large language models},
  author={Wang, Bailin and Wang, Zi and Wang, Xuezhi and Cao, Yuan and A Saurous, Rif and Kim, Yoon},
  journal={Advances in Neural Information Processing Systems},
  volume={36},
  pages={65030--65055},
  year={2023}
}

@inproceedings{
    evans2018can,
    title={Can Neural Networks Understand Logical Entailment?},
    author={Richard Evans and David Saxton and David Amos and Pushmeet Kohli and Edward Grefenstette},
    booktitle={International Conference on Learning Representations},
    year={2018},
    url={https://openreview.net/forum?id=SkZxCk-0Z},
}

@inproceedings{bowman-etal-2015-recursive,
    title = "Recursive Neural Networks Can Learn Logical Semantics",
    author = "Bowman, Samuel R.  and
      Potts, Christopher  and
      Manning, Christopher D.",
    editor = "Allauzen, Alexandre  and
      Grefenstette, Edward  and
      Hermann, Karl Moritz  and
      Larochelle, Hugo  and
      Yih, Scott Wen-tau",
    booktitle = "Proceedings of the 3rd Workshop on Continuous Vector Space Models and their Compositionality",
    month = jul,
    year = "2015",
    address = "Beijing, China",
    publisher = "Association for Computational Linguistics",
    url = "https://aclanthology.org/W15-4002/",
    doi = "10.18653/v1/W15-4002",
    pages = "12--21"
}

@inproceedings{rocktaschel2016reasoning,
    month={February}, 
    year={2016}, 
    author={Rocktaschel, Tim and Grefenstette, Edward and Hermann, Karl Moritz and Kocisky, Tomas and Blunsom, Phil}, 
    booktitle={International Conference on Learning Representations (ICLR)}, 
    title={Reasoning about Entailment with Neural Attention}, 
}

@article{cython,
  title={Cython: The best of both worlds},
  author={Behnel, Stefan and Bradshaw, Robert and Citro, Craig and Dalcin, Lisandro and Seljebotn, Dag Sverre and Smith, Kurt},
  journal={Computing in Science \& Engineering},
  volume={13},
  number={2},
  pages={31--39},
  year={2010},
  publisher={IEEE}
}

@inproceedings{z3,
author = {De Moura, Leonardo and Bj\o{}rner, Nikolaj},
title = {Z3: an efficient SMT solver},
year = {2008},
isbn = {3540787992},
publisher = {Springer-Verlag},
address = {Berlin, Heidelberg},
booktitle = {Proceedings of the Theory and Practice of Software, 14th International Conference on Tools and Algorithms for the Construction and Analysis of Systems},
pages = {337–340},
numpages = {4},
location = {Budapest, Hungary},
series = {TACAS'08/ETAPS'08}
}

@inproceedings{smt,
  title={The smt-lib standard: Version 2.0},
  author={Barrett, Clark and Stump, Aaron and Tinelli, Cesare and others},
  booktitle={Proceedings of the 8th international workshop on satisfiability modulo theories (Edinburgh, UK)},
  volume={13},
  pages={14},
  year={2010}
}

@unpublished{prover9,
  author = {W. McCune},
  title = {Prover9 and Mace4},
  note = {\url{http://www.cs.unm.edu/~mccune/prover9/}},
  year = {2005--2010}
}

@book{fol,
  title={A mathematical introduction to logic},
  author={Enderton, Herbert B},
  year={2001},
  publisher={Elsevier}
}

@book{ast,
  title={Compilers principles, techniques \& tools},
  author={Alfred, V Aho and Monica, S Lam and Jeffrey, D Ullman},
  year={2007},
  publisher={pearson Education}
}

@article{rbp1,
  title={Transition network grammars for natural language analysis},
  author={Woods, William A},
  journal={Communications of the ACM},
  volume={13},
  number={10},
  pages={591--606},
  year={1970},
  publisher={ACM New York, NY, USA}
}

@phdthesis{rbp2,
  title={A theory of syntactic recognition for natural language.},
  author={Marcus, Mitchell Philip},
  year={1978},
  school={Massachusetts Institute of Technology}
}

@inproceedings{mlp1,
  title={Constituency Parsing with a Self-Attentive Encoder},
  author={Kitaev, Nikita and Klein, Dan},
  booktitle={Proceedings of the 56th Annual Meeting of the Association for Computational Linguistics (Volume 1: Long Papers)},
  pages={2676--2686},
  year={2018}
}

@inproceedings{mlp2,
    title = "Multilingual Constituency Parsing with Self-Attention and Pre-Training",
    author = "Kitaev, Nikita  and
      Cao, Steven  and
      Klein, Dan",
    editor = "Korhonen, Anna  and
      Traum, David  and
      M{\`a}rquez, Llu{\'i}s",
    booktitle = "Proceedings of the 57th Annual Meeting of the Association for Computational Linguistics",
    month = jul,
    year = "2019",
    address = "Florence, Italy",
    publisher = "Association for Computational Linguistics",
    url = "https://aclanthology.org/P19-1340/",
    doi = "10.18653/v1/P19-1340",
    pages = "3499--3505"
}

@inproceedings{icl,
 author = {Brown, Tom and Mann, Benjamin and Ryder, Nick and Subbiah, Melanie and Kaplan, Jared D and Dhariwal, Prafulla and Neelakantan, Arvind and Shyam, Pranav and Sastry, Girish and Askell, Amanda and Agarwal, Sandhini and Herbert-Voss, Ariel and Krueger, Gretchen and Henighan, Tom and Child, Rewon and Ramesh, Aditya and Ziegler, Daniel and Wu, Jeffrey and Winter, Clemens and Hesse, Chris and Chen, Mark and Sigler, Eric and Litwin, Mateusz and Gray, Scott and Chess, Benjamin and Clark, Jack and Berner, Christopher and McCandlish, Sam and Radford, Alec and Sutskever, Ilya and Amodei, Dario},
 booktitle = {Advances in Neural Information Processing Systems},
 editor = {H. Larochelle and M. Ranzato and R. Hadsell and M.F. Balcan and H. Lin},
 pages = {1877--1901},
 publisher = {Curran Associates, Inc.},
 title = {Language Models are Few-Shot Learners},
 url = {https://proceedings.neurips.cc/paper_files/paper/2020/file/1457c0d6bfcb4967418bfb8ac142f64a-Paper.pdf},
 volume = {33},
 year = {2020}
}

@inproceedings{linc,
    title = "{LINC}: A Neurosymbolic Approach for Logical Reasoning by Combining Language Models with First-Order Logic Provers",
    author = "Olausson, Theo  and
      Gu, Alex  and
      Lipkin, Ben  and
      Zhang, Cedegao  and
      Solar-Lezama, Armando  and
      Tenenbaum, Joshua  and
      Levy, Roger",
    editor = "Bouamor, Houda  and
      Pino, Juan  and
      Bali, Kalika",
    booktitle = "Proceedings of the 2023 Conference on Empirical Methods in Natural Language Processing",
    month = dec,
    year = "2023",
    address = "Singapore",
    publisher = "Association for Computational Linguistics",
    url = "https://aclanthology.org/2023.emnlp-main.313/",
    doi = "10.18653/v1/2023.emnlp-main.313",
    pages = "5153--5176"
}

@inproceedings{symbcot,
    title = "Faithful Logical Reasoning via Symbolic Chain-of-Thought",
    author = "Xu, Jundong  and
      Fei, Hao  and
      Pan, Liangming  and
      Liu, Qian  and
      Lee, Mong-Li  and
      Hsu, Wynne",
    editor = "Ku, Lun-Wei  and
      Martins, Andre  and
      Srikumar, Vivek",
    booktitle = "Proceedings of the 62nd Annual Meeting of the Association for Computational Linguistics (Volume 1: Long Papers)",
    month = aug,
    year = "2024",
    address = "Bangkok, Thailand",
    publisher = "Association for Computational Linguistics",
    url = "https://aclanthology.org/2024.acl-long.720/",
    doi = "10.18653/v1/2024.acl-long.720",
    pages = "13326--13365"
}

@inproceedings{fairr,
    title = "{F}ai{RR}: Faithful and Robust Deductive Reasoning over Natural Language",
    author = "Sanyal, Soumya  and
      Singh, Harman  and
      Ren, Xiang",
    editor = "Muresan, Smaranda  and
      Nakov, Preslav  and
      Villavicencio, Aline",
    booktitle = "Proceedings of the 60th Annual Meeting of the Association for Computational Linguistics (Volume 1: Long Papers)",
    month = may,
    year = "2022",
    address = "Dublin, Ireland",
    publisher = "Association for Computational Linguistics",
    url = "https://aclanthology.org/2022.acl-long.77/",
    doi = "10.18653/v1/2022.acl-long.77",
    pages = "1075--1093"
}

@inproceedings{proofwriter,
    title = "{P}roof{W}riter: Generating Implications, Proofs, and Abductive Statements over Natural Language",
    author = "Tafjord, Oyvind  and
      Dalvi, Bhavana  and
      Clark, Peter",
    editor = "Zong, Chengqing  and
      Xia, Fei  and
      Li, Wenjie  and
      Navigli, Roberto",
    booktitle = "Findings of the Association for Computational Linguistics: ACL-IJCNLP 2021",
    month = aug,
    year = "2021",
    address = "Online",
    publisher = "Association for Computational Linguistics",
    url = "https://aclanthology.org/2021.findings-acl.317/",
    doi = "10.18653/v1/2021.findings-acl.317",
    pages = "3621--3634"
}

@inproceedings{explanation_refiner,
    title = "Verification and Refinement of Natural Language Explanations through {LLM}-Symbolic Theorem Proving",
    author = "Quan, Xin  and
      Valentino, Marco  and
      Dennis, Louise A.  and
      Freitas, Andre",
    editor = "Al-Onaizan, Yaser  and
      Bansal, Mohit  and
      Chen, Yun-Nung",
    booktitle = "Proceedings of the 2024 Conference on Empirical Methods in Natural Language Processing",
    month = nov,
    year = "2024",
    address = "Miami, Florida, USA",
    publisher = "Association for Computational Linguistics",
    url = "https://aclanthology.org/2024.emnlp-main.172/",
    doi = "10.18653/v1/2024.emnlp-main.172",
    pages = "2933--2958"
}

@inproceedings{faithful_refiner,
    title = "Faithful and Robust {LLM}-Driven Theorem Proving for {NLI} Explanations",
    author = "Quan, Xin  and
      Valentino, Marco  and
      Dennis, Louise A.  and
      Freitas, Andre",
    editor = "Che, Wanxiang  and
      Nabende, Joyce  and
      Shutova, Ekaterina  and
      Pilehvar, Mohammad Taher",
    booktitle = "Proceedings of the 63rd Annual Meeting of the Association for Computational Linguistics (Volume 1: Long Papers)",
    month = jul,
    year = "2025",
    address = "Vienna, Austria",
    publisher = "Association for Computational Linguistics",
    url = "https://aclanthology.org/2025.acl-long.867/",
    doi = "10.18653/v1/2025.acl-long.867",
    pages = "17734--17755",
    ISBN = "979-8-89176-251-0"
}
